\documentclass{article}

\usepackage[nonatbib]{neurips_2024}
\usepackage{neurips_2024}
\usepackage{multirow}
\usepackage{longtable} % For creating tables that span multiple pages
\usepackage{booktabs}  % For improved table formatting
\usepackage{pifont}
\usepackage[table]{xcolor}
\usepackage{adjustbox}
\usepackage{colortbl}
\usepackage{amsmath}
\usepackage[utf8]{inputenc} % allow utf-8 input
\usepackage[T1]{fontenc}    % use 8-bit T1 fonts
\usepackage{hyperref}       % hyperlinks
\usepackage{url}            % simple URL typesetting
\usepackage{amsfonts}       % blackboard math symbols
\usepackage{nicefrac}       % compact symbols for 1/2, etc.
\usepackage{microtype}      % microtypography
\usepackage[backend=biber, sorting=none]{biblatex}
\usepackage{url,hyperref,lineno,microtype,subcaption}
\usepackage[onehalfspacing]{setspace}
\usepackage{amssymb}
\usepackage{textcomp}
\usepackage{amsfonts} 
\usepackage{float}
\usepackage{afterpage}
\usepackage{placeins}
\usepackage{pifont}
\usepackage{xcolor}
\usepackage{nicefrac}
\usepackage{pdflscape}
\usepackage{ragged2e}
\usepackage{array}
\usepackage{geometry}
\usepackage{caption}
\usepackage{subcaption}

\title{Advancing Stroke Risk Prediction Using a Multi-modal Foundation Model}

\author{
  Camille Delgrange \\
  Signal Processing Institute\\
  EPFL University\\
  Lausanne, Switzerland \\
  \texttt{camille.delgrange@alumni.epfl.ch} \\
  \And
  Olga Demler \\
  Brigham and Women’s Hospital\\
  Harvard Medical School\\
  Boston, Massachusetts, USA\\
  \texttt{odemler@bwh.harvard.edu} \\
  \And
  Samia Mora\\
  Brigham and Women's Hospital \\
  Harvard Medical School\\
  Boston, Massachusetts, USA \\
  \texttt{smora@bwh.harvard.edu} \\
  \And
  Bjoern Menze\\
  Department of Quantitative Biomedicine \\
  University of Zurich \\ 
  Zurich, Switzerland \\
  \texttt{bjoern.menze@uzh.ch} \\
  \And
  Ezequiel de la Rosa\thanks{Equal contribution}\\
  Department of Quantitative Biomedicine \\
  University of Zurich \\ 
  Zurich, Switzerland \\
  \texttt{ezequiel.delarosa@uzh.ch} \\
  \And
  Neda Davoudi\thanks{Equal contribution, corresponding author}\\
  ETH AI Center, Department of Computer Science\\
  Department of Quantitative Biomedicine \\
  University of Zurich \\ 
  Zurich, Switzerland\\
  \texttt{neda.davoudi@ai.ethz.ch} \\  
}

\begin{document}

\maketitle

\begin{abstract}
    Predicting stroke risk is a complex challenge that can be facilitated by integrating diverse clinically available data modalities. This study introduces a self-supervised multi-modal framework combining brain MRI and clinical records to improve stroke risk prediction prior to onset. By leveraging large unannotated clinical datasets, the framework captures complementary and synergistic information across image (brain MRI) and tabular (clinical records) data modalities, serving as a multi-modal foundation model. Our approach employs a contrastive learning framework integrating language-image pretraining with an image-tabular matching module, to better capture the multimodal characteristics of each patient. 
    We benchmark its performance against state-of-the-art uni-modal and multi-modal methods across tabular, image and combined data modalities under various model settings. The proposed multi-modal foundation model outperformed self-supervised tabular (image) methods by 2.1\% (2.8\%) in ROC-AUC and by 10.6\% (12.8\%) in balanced accuracy. Additionally, it showed a 7.6\% increase in accuracy compared to the best multi-modal supervised model. Using interpretable tools, our approach demonstrated better integration of tabular and image data, providing richer and more aligned embeddings. Moreover, interpretable model-derived heatmaps further underscored brain regions frequently associated in the literature with brain aging, stroke risk, and clinical outcomes. This robust self-supervised multi-modal foundation model surpasses state-of-the-art methods for stroke risk prediction and lays the groundwork for future studies integrating diverse data modalities to advance clinical predictive modeling. Our approach is the first to allow the prediction of stroke risk prior to onset by leveraging multimodal clinical data, and could support earlier identification of at-risk patients and facilitate more proactive clinical management. 
\end{abstract}

\section{Introduction}
Stroke ranks as the second leading cause of death worldwide, responsible for 11.6\% of all deaths globally in 2019 \cite{Pu2023, Feigin2021}. It often results in neurological damage and long-term disability in adults, imposing significant health and economic challenges \cite{Pu2023, Feigin2021}. Early identification of patients at highest risk for stroke is crucial to prevent such events and long-term poor outcomes \cite{Flora2019}. 
The complexity of stroke, driven by multiple risk factors, highlights the importance of integrating multi-modal data to improve prediction accuracy and anticipate treatment strategies \cite{durso2024probabilistic}. Among the various imaging modalities, Magnetic Resonance Imaging (MRI) is uniquely suited to identify brain biomarkers potentially indicative of future stroke events. With its high-resolution, non-invasive capabilities, MRI enables precise assessment of structural abnormalities and detailed visualization of the brain's vascular network, offering critical insights into underlying conditions that may contribute to stroke risk \cite{Hartwig2009}.

\textbf{Uni-modal predictive models}
Predictive models play a crucial role in healthcare, as early identification of patients at risk is essential to prevent irreversible brain and body damage and ensure that they receive optimal care. Initially, these models were developed using a single clinical modality (uni-modal) such as images or clinical information from patient health records. Previous works mainly use convolutional neural networks (CNN) that can leverage high-dimensional imaging information for diagnosing patients \cite{Zhang2023}. Yu et al. applied deep learning algorithms to extract meaningful imaging features in an increasing order of hierarchical complexity to make predictions of the infarct volume \cite{yu2020use}. Other models that use only clinical records, often assume linear relationships between traditional risk factors such as age, gender, smoking status, blood pressure, diabetes, cholesterol levels, and body mass index \cite{Hippisley-Cox2024, An2020, You2023}. Alaa et al. used AutoPrognosis, an ensemble machine learning approach, to outperform conventional models like the Framingham score and Cox models \cite{Alaa2019}. A major limitation of these models is their inability to integrate complementary information from other modalities, unlike clinicians who rely on multiple data sources for diagnosis. Biobanks like the UK Biobank (UKB) has become invaluable in this context, providing vast datasets integrating imaging and clinical information to train machine learning models for disease prediction \cite{Littlejohns2020, UKBioBank2021}.

\textbf{Multi-modal predictive models}
Several studies have integrated diverse data types to improve diagnostic capabilities \cite{Liu2015}. For example, de Havenon et al. \cite{deHav2023improvement} highlighted the value of imaging biomarkers by improving cerebrovascular events  prediction through the incorporation of white matter hyperintensities derived from brain imaging into clinical records, emphasizing the potential of such approaches to refine risk assessments. MultiSurv has shown success by fusing image and tabular data for cancer survival prediction \cite{Vale-Silva2021}. In another study, integration of retinal images and clinical records was leveraged to improve cardiovascular disease prediction \cite{huang2024predictingstrokeretinalgraphs}. Multi-modal models combining imaging data and clinical records have demonstrated better prediction performance for disability prediction in stroke patients \cite{White2024, Liu2023}.
However, CNNs tend to prioritize image features, and simple image-tabular CNN concatenation fails to enhance predictive models due to insufficient cross-modal interactions. To address this, Wolf et al. developed the Dynamic Affine Feature Map Transform (DAFT), which conditions convolutional feature maps on both image and tabular data, enabling a two-way information exchange via an auxiliary neural network \cite{Wolf2022}. While DAFT reduces issues related to the large number of trainable parameters in standard 3D CNNs and the curse of dimensionality, it may sacrifice some predictive power compared to deeper models like ResNet. Although recent models show promise in biomedical prediction tasks, their clinical translation is hindered by limited annotated datasets, low disease prevalence, and the risk of overfitting. Self-supervised learning (SSL) is a powerful technique for extracting representative features from unlabeled data, making it valuable for early disease risk identification.

\textbf{Self-supervised models}
Unlike traditional supervised learning, SSL defines pretext tasks allowing models to learn meaningful representations from raw data \cite{Balestriero2023}. One prominent SSL technique is contrastive learning, which trains encoders to generate augmented views of a sample, maximizing similarity between these views while minimizing similarity with other samples \cite{Balestriero2023}. Popular methods such as SimCLR \cite{Chen2020}, BYOL \cite{Grill2020}, and MOCO \cite{He2020} have demonstrated success in imaging tasks, while VIME \cite{Houthooft2016} and SCARF \cite{Bahri2022} are leading approaches for tabular data.
Emerging approaches, like contrastive language-image pre-training (CLIP) strategy, have evolved from uni-modal methods to integrate diverse modalities. While there was an extensive work done for cardiovascular diseases prediction \cite{Radhakrishnan2023, Du2024, Hager2023, girlanda2024enhancing}, stroke risk prediction through volumetric brain images and clinical health records remains underexplored. 

We present for the first time, to the best of our knowledge, a scalable self-supervised multi-modal approach integrating research quality T2-FLAIR MRIs pre-stroke, along clinical records for stroke risk prediction, prior to onset. As depicted in Figure \ref{fig:model_design}, our methodology incorporates cross-modal interactions via CLIP loss \cite{Radford2021} and image-tabular matching (ITM) loss \cite{Li2021, Du2024}. We demonstrate that our learning strategy outperforms leading (self-)supervised uni-modal methods and that multi-modal image-tabular pre-training aligns data modalities in a shared space improving representations, as well as  downstream task performance. Lastly, we validate the model learned features through visual activation maps, which align with established clinical and neurological findings on stroke-related brain pathology. Our model is practical for real-world clinical use because it leverages a small dataset without the need of expert annotations, and could facilitate early identification of at-risk patients for a more proactive clinical management. The code repository for this project is available at the following link: \url{https://github.com/CamilleDelgrange/SSMSRPM/}.

\section{Materials and Methods}
\label{sec:Materials_and_Methods}
\subsection{Dataset}
\label{subsec:dataset}
\subsubsection{UKB Dataset}
Our analyses are performed on the UKB, using a single (pre-stroke) T2-Fluid Attenuation Inversion Recovery (FLAIR) scan with timewise corresponding clinical patient information spanning across five categories: demographics, lifestyle, biomarkers, comorbidities, and medication. T2-FLAIR scans serve as an effective MRI sequence for identifying white matter hyperintensities (WMHs), which are strongly associated with increased stroke risk and serve as critical markers of cerebrovascular damage that may precede and predict stroke occurrence \cite{Shim2015, Debette2010, Liou2009, Leiva-Salinas2010}. Stroke patients were identified using International Classification of Disease (ICD10) codes specific to stroke \cite{Woodfield2015, Category43UKB} (ICD- 9 codes 430, 431, 434 and 436; ICD- 10 codes I60, I61, I63 and I64). ICD10 codes are maintained by the World Health Organization, used to record diagnoses during hospital admissions, and made available through the UKB. The stroke onset time was defined as the time between the baseline first imaging visit and the stroke onset. For healthy participants, they are defined as still healthy (e.g. without any stroke event) at the (second) repeated imaging timepoint. A summary of the cohort demographics including time-dependent variables can be found in Table \ref{tab:demographics}. The time variables were not used for training or inference however they give insights into the mean stroke onset time (from scan acquisition until the recorded clinical stroke event) as well as the last timepoint where the patient attended an imaging visit as healthy. The complete list of features used is available in Tables S1 and S2 in the supplementary materials.
Continuous features are standardized using z-score normalization, while categorical data are one-hot encoded. For our experiments, as indicated in Table \ref{tab:dataset_splits} during the model pre-training stage, we use 5000 and 500 samples for the training and validation sets, respectively. For the downstream fine-tuning stage, the training, validation, and test subsets -none of which were used during the pre-training stage- include 278, 93, and 93 samples, respectively. The fine-tuning sets are stratified by sex, age, and stroke diagnosis to address class imbalance; the splits were designed to preserve the distribution of these variables as observed in the pre-training dataset. 
To handle missing tabular data, we use an iterative multivariate imputer based on Multivariate Imputation by Chained Equations (MICE) \cite{VanBuuren2011}, modelling missing features as a function of existing features over multiple imputation rounds. Missing categorical data is replaced by the most frequent category. This step is performed after data normalization, to ensure that the means and standard deviations are calculated only from recorded values. 
\subsubsection{Brain MRI pre-processing}
The 3D brain MRI images are registered to the Montreal Neurological Institute (MNI) brain template space \cite{MNI}, standardized to dimensions of $182 \times 218 \times 182$ voxels with a voxel size of $1~\mathrm{mm}^3$, and processed using the UK Biobank imaging pipeline \cite{Alfaro-Almagro2018}. Key image-derived phenotypes (IDPs), such as brain tissue and white matter hyperintensity (WMH) volumes are extracted and referred to as “brain IDPs.” Segmentation of pre-stroke brain lesion abnormalities is performed using the BIANCA tool, generating 3D binary lesion masks \cite{Alfaro-Almagro2018}. Additionally, features such as volume, area, and elongation are extracted from the lesion segmentation masks using pyradiomics \cite{van2017computational}; this set of features is referred to as “lesion IDPs” throughout the paper.

\FloatBarrier
\begin{longtable}[H]{>{\centering\arraybackslash}p{2.7cm} 
                  >{\centering\arraybackslash}p{2.7cm} 
                  >{\centering\arraybackslash}p{2.5cm} 
                  >{\centering\arraybackslash}p{2.5cm}
                  >{\centering\arraybackslash}p{1.5cm}} % Center-align all columns
\captionsetup{labelformat=empty} % Suppress the automatic "Table 2" generation
\caption{\textbf{Table 1.} Characteristics of the inference cohort for healthy individuals and stroke patients.}
\label{tab:demographics}
\\
\hline
\textbf{Category} & \textbf{Feature (unit)} & \textbf{Stroke} & \textbf{Healthy} & \textbf{$p$-value} \\
\hline
\endfirsthead

\hline
\textbf{Category} & \textbf{Feature (unit)} & \textbf{Stroke} & \textbf{Healthy} & \textbf{$p$-value} \\
\hline
\endhead

\hline
\centering \textbf{Demographics} 
& Sex (Male, \%) & 60.87\% & 51.06\% & 0.4572\\
& Age (years) & 67.70 ± 7.30 & 61.38 ± 9.68 & 0.0006\\
& BMI ($kg/m^{2}$) & 28.01 ± 5.00 & 25.79 ± 3.82 & 0.0211\\
& Waist Circumference (cm) & 93.34 ± 14.37 & 86.68 ± 13.14 & 0.0247\\
& Weight (kg) & 80.86 ± 16.76 & 74.94 ± 15.21 & 0.7310\\
\hline
\centering \textbf{Time Variables} 
& Stroke Onset (months) & 26.20 ± 18.79 & N/A \\
& Healthy Window (months) & N/A & 29.93 ± 10.30 \\
\hline
\end{longtable}

\begin{table}[H]
    \centering
    \caption{Summary of dataset splits used during the pre-training and fine-tuning stages.}
    \label{tab:dataset_splits}
    \begin{tabular}{lcc}
        \\
        \toprule
        \textbf{Stage} & \textbf{Subset} & \textbf{Number of Samples} \\
        \midrule
        \multirow{2}{*}{Pre-training} 
            & Training   & 5000 \\
            & Validation & 500 \\
        \midrule
        \multirow{3}{*}{Fine-tuning} 
            & Training   & 278 \\
            & Validation & 93  \\
            & Test       & 93  \\
        \bottomrule
    \end{tabular}
\end{table}

\subsection{Multi-modal self-supervised framework}
\label{subsec:SSL_framework}
Our pipeline is split into two sequential steps. First, we pre-train the tabular and imaging encoders (Figure \ref{fig:model_design} A) and then we fine-tune them with labels from the downstream task (Figure \ref{fig:model_design} B). The image encoder consists of a ResNet-50 backbone \cite{He2020}, which outputs feature embeddings of size 512. The tabular encoder is a multi-layer perceptron (MLP) composed of two fully connected layers, each with an embedding dimension of 512 neurons. Each layer is followed by batch normalization, ReLU activation, and dropout with a rate of 0.3. Kaiming weight initialization is applied, and pretrained weights can optionally be loaded and frozen. The MLP encodes structured tabular data into dense embeddings. Both encoders are followed by a projection head, as used in SimCLR \cite{Chen2020}, which maps the embeddings to a lower-dimensional space for contrastive learning (128). The projection head consists of a fully connected hidden layer with ReLU activation, followed by a second linear layer without activation. This two-layer MLP allows the model to learn more expressive representations in the contrastive space, decoupled from the original embedding space. 
Each batch of data contains pairs of imaging $x_{j_i}$ and tabular $x_{j_t}$ samples which are augmented by random transformations \( t \sim \tau \) from a set of parametric transforms \(\tau\), such as random cropping and affine transforms for the images, or random feature corruption for the tabular data. We use an image augmentation rate of 95\% for the model to still occasionally see unaltered data to capture the original data distribution for transfering the learnt features to the downstream task. The corruption rate of the tabular data is set to 0.3 as in the original SCARF method \cite{Bahri2022}. 
\subsubsection{Model pre-training}
For a given reference point, known as anchor \( x \), the positive samples are the ones derived from \( x \) with transformations while other samples in the batch are considered as negative samples. 
Augmented images $x_{j_i}$ and tabular data $x_{j_t}$ are passed through the imaging $f_{\theta_I}$ and tabular encoder $f_{\theta_T}$ to generate the embeddings.  
These embeddings are propagated through the separate projection heads $f_{\phi_I}$ and $f_{\phi_T}$, and brought into a shared latent space as projections $z_{j_i}$ and $z_{j_t}$, which are l2-normalized onto a unit hypersphere. The projections are pulled and pushed in the shared latent space according to the CLIP loss \cite{Radford2021}, which maximizes the cosine similarity of projections from the positive samples and minimizes the similarity of projections from the negative samples in the batch. In contrast to the original InfoNCE loss used in SimCLR \cite{Chen2020}, and following the CLIP loss, the projected embeddings similarities are contrasted between data modalities. An image projection is therefore defined as in Equation \ref{eq:01}:
\begin{equation}
    z_{j_i} = f{\phi_I}(f_{\theta_I}(x_{j_i}))\label{eq:01}
\end{equation}
Considering all N subjects in a batch, the loss for the imaging modality $l_{i,t}$ is defined as follows, in Equation \ref{eq:02}:
\begin{equation}
    l_{i,t} = -\sum_{j \in N}^{}log\frac{exp(cos(z_{j_i}, z_{j_t})/\tau)))}{\sum_{k\in N, k\neq j}^{}exp(cos(z_{j_i}, z_{k_t})/\tau)))}\label{eq:02}
\end{equation}
where $\tau$ is the temperature parameter. In our experiments, a temperature of 0.1 is selected to work best, following \cite{Chen2020}.
$l_{t,i}$ is computed analogously and the CLIP loss is defined in Equation \ref{eq:03} as follows:
\begin{equation}
    \mathcal{L}_{clip} = \lambda l_{i,t} + (1-\lambda)l_{t,i}\label{eq:03}
\end{equation}

\justify We choose a value of 0.5 for the regularization parameter $\lambda$. The goal is to learn patient-specific representations that remain consistent despite variations in the image-tabular pairs. 
Hard negative samples are crucial in contrastive learning as they help the model to distinguish between similar samples, preventing trivial solutions and enhancing its robustness. We implement a hard negative mining strategy to predict whether image-tabular data pairs are positive or negative, using the image-tabular matching (ITM) loss. In this approach, for each image or tabular representation, we identify an unmatched tabular or image representation from the mini-batch. This selection is based on similarity scores computed using the CLIP method, which serves as the sampling weight for the negative pairs \cite{Du2024, Li2021}. A multi-modal interaction module is introduced, as shown in Figure \ref{fig:model_design}, which takes the output of the projector heads to perform inter-modality learning and generates a multi-modal representation. It uses a cross-attention mechanism \cite{Vaswani2017}, enabling tabular embeddings to attend to relevant image embeddings. 
The multi-modal interaction module contains two transformer layers, with four attention heads and a hidden dimension of 256, each including self-attention, cross-modal attention, an MLP feed-forward module and layer normalization \cite{Du2024}. 
The output of the multi-modal module is a classification token [CLS], typically placed at the start of the input sequence to represent aggregated contextual information, used for downstream classification task, where the model needs a single feature vector representing the entire input \cite{Oh2017}. The [CLS] embedding is capturing a joint representation of the image-tabular pair that is fed into the ITM predictor (namely, a linear layer) to match the prediction based on a binary cross-entropy loss $\mathcal{L}_{ITM}$. 
Therefore, the complete loss $\mathcal{L}$ is expressed in Equation \ref{eq:04}, as: 
\begin{equation}
    \mathcal{L} = (\mathcal{L}_{CLIP} + \mathcal{L}_{ITM})/2\label{eq:04}
\end{equation}

\subsubsection{Model fine-tuning: downstream task predictions}
\begin{figure}[htbp!]
  \begin{center}
  \includegraphics[width=1.3\textwidth]{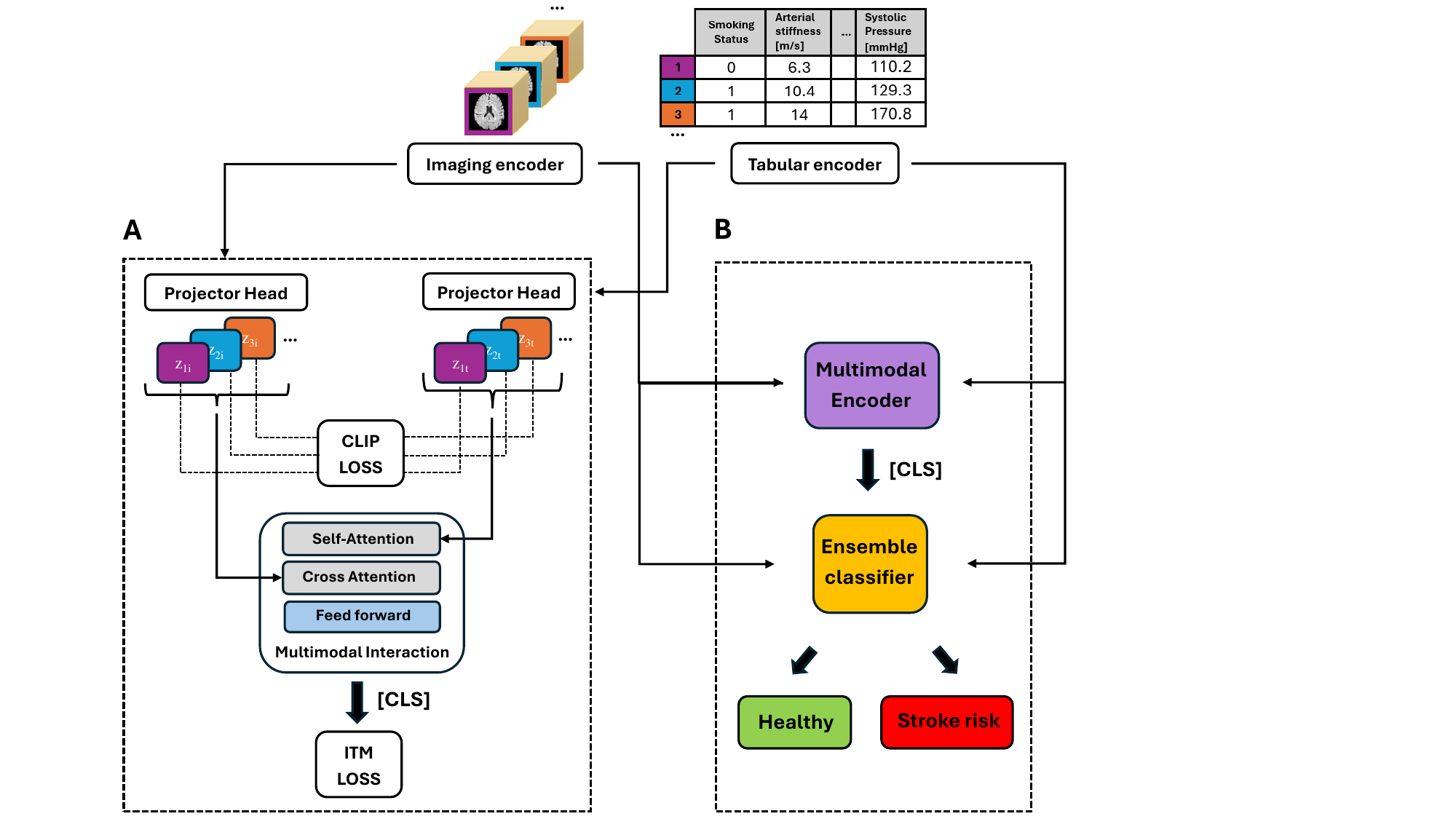}
  \end{center}
  \caption[Overview of the method.] {Pipeline for joint imaging and tabular data pre-training (A) and supervised fine tuning (B). CLIP loss is applied on projected data to align the image and tabular representations. Hard negative pairs are mined through CLIP similarities within the batch. A transformer block with self-attention and cross-attention layers is used to cross-attend both modalities, resulting in a multi-modal [CLS] token fed to a classifier and used for further downstream fine tuning. Image-Tabular Matching (ITM) loss evaluates the image-tabular pair matching. In the downstream task, an ensemble classifier is fine-tuned to predict healthy versus stroke from pre-trained imaging, tabular and multi-modal encoders.}
  \label{fig:model_design}
\end{figure}

After pre-training, the projection heads are replaced by fully connected layers. Extracting the representation before the projector has been shown to improve downstream tasks performance \cite{Balestriero2023}. For downstream fine-tuning and binary classification of samples into healthy \emph{versus} stroke (Figure \ref{fig:model_design} B), we employ ensemble learning to improve model generalization and performance by leveraging the rich representations from the image encoder, tabular encoder, and the multi-modal transformer interaction module. 
All pre-trained models are evaluated using linear probing (frozen) and fine-tuning (trainable). The frozen models use tuned linear classifiers after the feature extractors. 
The datasets used for model fine-tuning are balanced in each batch of training, validation, and test subset. This way, we reduce potential bias due to class-imbalance, as well as unstable and slow training due to imbalance batch distributions. We report test set results in Table \ref{tab:performance_metrics}. However, we also report test set results reflecting the original (imbalanced) stroke distribution (0.52\% in the original UKB pre-training dataset), using a bootstrapping with replacement approach. These results are presented in Table \ref{tab:bootstrapped_results} and are representative of the distribution in the UKB dataset used in this study. Additionally, we provide 95\% confidence intervals derived from this analysis.

\section{Experiments} \label{sec:Experiments}
\subsection{Benchmarking}
\label{subsec:benchmarking_exp}
The herein proposed solution is compared against supervised and SSL strategies, each of them using imaging, tabular, and integrated imaging-tabular methodologies.

\subsubsection{Supervised learning methods}
\label{subsubsec:Sup_methods}
To benchmark our proposed method, we implement two state-of-the-art, supervised image-based models, namely ResNet50 \cite{He2020} and DenseNet121 \cite{Huang2017}, two supervised tabular data approaches, namely a tabular MLP model and a tabular transformer encoder inspired by Du et al. (2024) \cite{Du2024}. We conduct an ablation study using a supervised MLP model with various feature combinations to identify the optimal feature set. This process helps us to select the final combination of features for improved model performance. The combinations include: $i)$ clinical records only, spanning the previously mentioned categories (referred to as "clinical"), $ii)$ clinical records with brain extracted IDPs ("clinical + brain IDPs"), and $iii)$ clinical records, brain IDPs and lesion IDPs ("clinical + brain IDPs + lesion IDPs"). Furthermore, we implement three supervised, multi-modal (imaging-tabular) learning models, namely a simple concatenation fusion model (CF) \cite{Spasov2019}, a CF model integrated with the tabular transformer encoder inspired by the work from Du et al. (2024) \cite{Du2024} (CF + Transformer), %instead of the tabular MLP encoder, 
and DAFT model \cite{Wolf2022}. All models employing an imaging encoder are implemented with ResNet50 as a backbone. The DAFT block is integrated within ResNet50 from the third stage onward. To alleviate overfitting, an early stopping strategy is adopted, with a minimal delta (divergence threshold) of $1 \times 10^{-4}$, a maximal number of epochs of 50, and a patience of 15 epochs.

\subsubsection{Self-supervised learning methods}
\label{subsubsec:SSL_methods}
Our model is compared against leading, self-supervised contrastive solutions, including: $i)$ the uni-modal, image-based SimCLR\cite{Chen2020} approach, $ii)$ the uni-modal, tabular data-based SCARF \cite{Bahri2022} approach, and $iii)$ the multi-modal, CLIP method (without ITM loss). The hyperparameters and training configurations for all SSL pre-training approaches are adapted for our specific dataset and task, and are obtained through hyper-parameter search. All models are pre-trained for 100 epochs using an Adam optimizer \cite{Kingma2015}. The learning rate is warmed up linearly for 10 epochs and decayed following a cosine annealing scheduler. For all methods, the image augmentation rate is 95\% and the tabular corruption rate is 0.3 during pre-training and 80\% and 0.3 during supervised fine tuning. SimCLR is trained using the NTXent objective \cite{Chen2020} and the temperature parameter is kept to 0.1. Hidden and projected dimensions are respectively 2048 and 128 for both modalities \cite{Chen2020}. The same parameters are used for SCARF \cite{Bahri2022}. Learning rates are chosen with a sweep through a range of learning rates, by tracking the validation loss. Weight decay and dropout rate are added depending on the level of overfitting observed at the validation loss. The same early stopping strategy is employed as in the supervised learning methods.
The downstream fine-tuning employs the same parameters as the pre-training method, using a single modality classifier for the unimodal SSL pre-trained methods and a fused representation vector combined with a single linear classifier for the CLIP-only model. Trainable models in each SSL uni-modal method means that the other modality is incorporated during fine-tuning as a full trainable model. 
The employed batch size for all methods is 6. All SSL models are pretrained on a Tesla V100-SXM2 (32GB, 42 CPUs), and inference is performed on an NVIDIA GeForce RTX 4090 (24GB, 62 CPUs). The pretraining lasts around $\sim$ 24 hours, while fine-tuning less than 1 hour.
Tables detailing the training parameters used during both pre-training and fine-tuning are included as Tables S4 and S5 in the Supplementary Material.

\subsection{Interpretability and qualitative analysis}
\label{subsec:interpretability_exp}
Embeddings visualization is done using a two-dimension Uniform Manifold Approximation and Projection (UMAP) technique \cite{McInnes2020}, to evaluate the quality of the generated latent space embedding after pre-training, using validation samples. Such approach allows to qualitatively assess the latent space representation and data distribution after encoding each modality with either a uni-modal or multi-modal pre-trained model, providing insights into the effectiveness of the learning strategies. A latent space embedding size of 2048 dimensions is produced.

For the qualitative analysis, we are interpreting the model's predictions with respect to both MRI and clinical features modality. First, to analyze the importance of the individual tabular features in generating the embeddings, we compute the integrated gradients over the test samples with a multi-modal pre-trained tabular encoder, aggregating the importance of each feature across all embeddings dimensions, specifically handling categorical features by summing the mean importance over all choices. Integrated gradients are computed with the package Captum integrated with pytorch \cite{kokhlikyan2020captum}. Then, we generate imaging heatmaps using the Gradient-weighted Class Activation Mapping (GradCAM) technique \cite{Selvaraju2017} to visualize the regions in each slice that contributed most significantly to the model's predictions across the brain MRI volumes.
GradCAM \cite{Selvaraju2017} heatmaps are normalized in the range of 0 to 1 and are upsampled with trilinear interpolation to match the original image space. The 7th layer of the ResNet50 encoder is used to allow capturing high-level features and spatial structure that is suitable for visualization. The most informative slice (defined as the slice with the highest heatmap activation scores) for each view in axial, sagittal, or coronal plane is generated. We use the 3D GradCAM implementation from MONAI \cite{Cardoso2022}. 
Pursuing our analysis, we generate boxplots comparing the normalized mean GradCAM activations for healthy versus stroke individuals on  WMH-segmented brain MRIs and on whole T2-FLAIR brain MRI scans.

\subsection{Performance assessment}
\label{subsec:perf}
All models are evaluated through the area under the Receiver Operating Characteristic (ROC) curve. Binary classification metrics, namely (balanced) accuracy, F1-Score, and sensitivity, are included. Classification metrics are reported at the Youden-index operating point ($J = \text{Sensitivity} + \text{Specificity} - 1$) retrieved from the (validation set) ROC curve. The metrics are chosen bearing in mind that potential clinical applications of this study could serve as screening and risk stratification tools, where the model's sensitivity plays an important role to avoid missing positive stroke cases. 

\section{Results and Discussion}
\label{sec:results_discussion}
\subsection{Benchmarking}
\label{subsec:benchmarking_results}
To determine which tabular features to include, we conducted a supervised-learning ablation analysis using various combinations of tabular data subgroups. As shown in Table \ref{tab:ablation_study_features_selection}, the models that incorporate clinical and brain IDPs achieve the highest ROC-AUC scores. However, the method that also includes lesion IDPs outperforms in binary classification metrics, such as F1-score and sensitivity. To prioritize model robustness while maintaining a smaller feature set, the subsequent benchmarking of models using tabular data is performed using only clinical and brain IDPs.

\begin{table}[H]
    \centering
    \caption{Feature selection with supervised MLP using different combinations of tabular data. ROC-AUC: area under the receiver operating characteristic curve; Acc (\%): Accuracy; Se (\%): sensitivity. For each metric, the best-performing method is highlighted in \textbf{bold} and the second-best is \underline{underlined}.}    \label{tab:ablation_study_features_selection}
    \vspace{0.5cm} % Add vertical space (adjust the value as needed)
\makebox[\textwidth][c]{ % Forces the table to be centered
\begin{tabular}{lcccccc}
    \toprule
    \multirow{2}{*}{Model} & \multicolumn{4}{c}{Metrics (\%)} \\
    \cmidrule(r){2-5} & \textbf{AUC} & \textbf{bAcc} & \textbf{F1} & \textbf{Se} \\
    \midrule
    MLP (clinical) & 65.36 & \underline{61.29} & 60.87 & 60.87 \\
    MLP (clinical + brain IDPs) & \textbf{71.37} & 60.31 & \underline{65.96} & \underline{67.39} \\
    MLP (clinical + brain IDPs + lesion IDPs) & \underline{65.63} & \textbf{62.58} & \textbf{68.69} & \textbf{73.91} \\
    \bottomrule
    \end{tabular}}
\end{table}

A summary of the different models performance is shown in Table \ref{tab:performance_metrics}. It is observed that the proposed multi-modal learning strategy outperforms all other methodologies across all considered metrics, with the \emph{trainable} model setting performing slightly better than the \emph{frozen} one.

\begin{table}[H]
    \centering
    \caption{Benchmarking performance results. F and T denote frozen and trainable pre-trained encoders, respectively. ROC-AUC: area under the receiver operating characteristic curve; Acc (\%): Accuracy; Se (\%): Sensitivity. For each metric, the best-performing method is highlighted in \textbf{bold} and the second-best is \underline{underlined}. The overall best performing method is highlighted in gray.}
    \label{tab:performance_metrics}
    \vspace{0.5cm} % Add vertical space (adjust the value as needed)
\makebox[\textwidth][c]{ % Forces the table to be centered
\begin{tabular}{lcccccc}
    \toprule
    \multirow{2}{*}{Model} & \multicolumn{1}{c}{Tabular} & \multicolumn{1}{c}{Image} & \multicolumn{4}{c}{Metrics (\%)} \\
    \cmidrule(r){2-3} \cmidrule(r){4-5} \cmidrule(r){4-7} & \textbf{} & \textbf{} & \textbf{AUC} & \textbf{Acc} & \textbf{F1} & \textbf{Se}  \\
    \midrule
    \multicolumn{7}{c}{(a) Supervised Image} \\
    \midrule
    ResNet-50 \cite{He2016} & - & T & 63.25 & 57.08 & 60.01 & 65.22 \\
    DenseNet121 \cite{Huang2017} & - & T & 66.79 & 66.79 & 69.90 & 78.26 \\
    \midrule
    \multicolumn{7}{c}{(b) Supervised Tabular} \\
    \midrule
    MLP & T & - & 71.37 & 60.31 & 65.96 & 67.39 \\
    Transformer \cite{Du2024} & T & - & 64.38 & 62.21 & 47.62 & 32.61 \\
    \midrule
    \multicolumn{7}{c}{(c) Supervised multi-modal} \\
    \midrule
    Concat Fuse (CF) \cite{Spasov2019} & T & T & 65.26 & 60.29 & 62.63 & 67.39 \\
    Concat Fuse (CF) [w/ Transformer] & T & T & 63.48 & 52.08 & 66.17 & 95.65 \\
    DAFT \cite{Wolf2022} & T & T & 73.82 & 63.51 & 69.57 & 65.01 \\
    \midrule
    \multicolumn{7}{c}{(d) SSL Image} \\
    \midrule
    SimCLR \cite{Chen2020} & - & F & 64.99 & 52.38 & 33.33 & 28.91  \\
    SimCLR \cite{Chen2020} & - & T & 65.59 & 55.67 & 43.83 & 34.78  \\
    SimCLR \cite{Chen2020} & T & F & 72.02 & 65.56 & 64.44 & 63.04 \\
    SimCLR \cite{Chen2020} & T & T & 72.11 & 65.56 & 64.44 & 63.04 \\
    \midrule
    \multicolumn{7}{c}{(e) SSL Tabular} \\
    \midrule
    SCARF \cite{Bahri2022} & F & - & 71.18 & 62.42 & 63.91 & 67.39 \\
    SCARF \cite{Bahri2022} & T & - & 70.35 & 64.48 & 62.92 & 60.87 \\
    SCARF \cite{Bahri2022} & F & T & 72.16 & 62.16 & 43.48 & 53.34 \\
    SCARF \cite{Bahri2022} & T & T & 72.02 & 67.85 & \underline{73.08} & 78.26 \\
    \midrule
    \multicolumn{7}{c}{(f) SSL multi-modal} \\
    \midrule
    CLIP \cite{Hager2023} & T & T & 73.41 & 61.5 & 67.24 & \underline{80.78} \\
    CLIP \cite{Hager2023} & F & F & 73.54 & \underline{71.00} & 70.97 & 71.74 \\
    \rowcolor{lightgray} Ours & T & T & \underline{74.42} & \textbf{71.11} & \textbf{74.22} & \textbf{84.78} \\
    \rowcolor{lightgray} Ours & F & F & \textbf{74.75} & 62.77 & 67.29 & 76.60 \\
    \bottomrule
    \end{tabular}}
\end{table}
\FloatBarrier

When comparing models based on learning approach and data modality, it can be observed that the best performing imaging supervised learning strategy is DenseNet121 (ROC-AUC 66.79\%). In DenseNet architectures, layers are densely connected, which improves feature reuse and gradient flow, leading to richer feature representations. However, this dense connectivity increases memory overhead during training, particularly with the large inputs used in this study. To optimize the trade-off between efficiency and memory usage, we selected ResNet50 as the encoder for SSL pre-training, accepting a minor reduction in performance.

When comparing SSL strategies, it is evident that fine-tuning both data modalities in multi-modal approaches significantly boosts performance. The performance gap is considerable when comparing these multi-modal models with uni-modal image-based models, showing that image data alone is insufficient for effectively addressing the task.
Our model is performing better than all uni-modal (tabular and imaging) SSL methods, as well as the method using only CLIP loss. Interestingly, DAFT performs similarly to the multi-modal SSL methods in terms of ROC-AUC and accuracy, although exhibiting poor F1-score and sensitivity results. There is no clear difference in performance between trainable and frozen settings across all models. We hypothesize this is because the pre-trained models have already developed robust, transferable representations, making fine-tuning less impactful. Additionally, the small size of the fine-tuning dataset may limit the effectiveness of further learning beyond what was achieved during pre-training. Besides, it could be hypothesized that freezing the model may serve as a form of regularization, helping to mitigate overfitting, particularly in this setting with limited labeled data.

To assess statistical significance of our model's performance superiority on the true stroke prevalence from the original pre-training stroke distribution, we have performed Wilcoxon signed-rank tests comparing each model against ours in Table \ref{tab:bootstrapped_results}.The results indicate that our model significantly outperforms all other uni-modal and multi-modal unsupervised baselines, with $p$-values~$<~0.01$.

\begin{table}[ht]
\centering
\caption{Test set performance of unsupervised models for stroke classification. Results were obtained via bootstrapping with replacement (1,000 iterations), preserving the real-world stroke prevalence in the UK Biobank (~0.52 \%). Statistical significance was assessed using the Wilcoxon signed-rank test, comparing the baseline models against our proposed approach. AUC: area under the receiver operating characteristic curve. BA: balanced accuracy. CI: confidence intervals. \textasteriskcentered\ denotes $p$-values~$<~0.01$.}
\label{tab:bootstrapped_results}
\vspace{0.5cm} % Add vertical space (adjust the value as needed)
\resizebox{\textwidth}{!}{%
\begin{tabular}{lcccc}
\toprule
\textbf{Model} & \textbf{AUC [95\% CI]} & \textbf{BA [95\% CI]} & \textbf{F1-score [95\% CI]} & \textbf{Sensitivity [95\% CI]} \\
\midrule
SimCLR & 0.720 [0.667--0.772]\textasteriskcentered & 0.593 [0.556--0.628]\textasteriskcentered & 0.6456 [0.6213--0.7362]\textasteriskcentered & 0.631 [0.530--0.730]\textasteriskcentered \\
SCARF & 0.727 [0.672--0.780]\textasteriskcentered & 0.615 [0.571--0.659]\textasteriskcentered & 0.7108 [0.7024--0.7229]\textasteriskcentered & 0.740 [0.650--0.830]\textasteriskcentered \\
CLIP & 0.734 [0.680--0.787]\textasteriskcentered & 0.656 [0.606--0.705]\textasteriskcentered & 0.7127 [0.6983--0.7342]\textasteriskcentered & 0.782 [0.700--0.860]\textasteriskcentered \\
\rowcolor{lightgray} \textbf{Ours} & \textbf{0.748 [0.696--0.798]} & \textbf{0.721 [0.677--0.761]} & \textbf{0.7443 [0.7233--0.7652]} & \textbf{0.846 [0.770--0.910]} \\
\bottomrule
\end{tabular}}
\end{table}

In terms of model performance with respect to the stroke onset time variable, shown in Figure \ref{fig:stroke_onset_time}, the highest accuracy is achieved for people developing stroke after 2-3 years. First, strokes occurring closer to the scan acquisition date might be driven by acute mechanisms, such as embolism from an undiagnosed cardiac condition or sudden plaque rupture \cite{Adams1993}. These acute events may not leave detectable brain MRI changes prior to stroke onset. Second, the model might be biased towards more chronic changes, long-term stroke risk factors as the dataset is skewed towards longer timeframes. However, the small sample size (N=93), might also be insufficient to draw such conclusions. Further, in Figure \ref{fig:age}, the high proportion of correctly predicted strokes in older age groups (particularly 70-80), as well as the higher F1 score suggests that age is not only the strongest predictor variable, but it may also correlate with features that the model recognizes more confidently as indicative of stroke. In the healthy group, this trend is less pronounced, indicating that the model’s outputs are less influenced by the individuals’ ages. Given the inherent relationship between age and stroke risk, we conducted a logistic regression analysis to ensure the model is not undesirably inferring age as a proxy for stroke risk rather than predicting an individual's true stroke risk. This analysis included various combinations of predictor variables, such as patient age, total WMH volume in brain regions (a known mediator of the relationship between age and stroke risk), and the predicted probabilities from the best-performing model from Table \ref{tab:performance_metrics}. As shown in Table S3 in the supplementary materials, incorporating the predicted probabilities into the logistic regression model yields approximately a 20\% improvement compared to models based solely on age or WMH volume.
Age and WMH volume are complementary predictors: when combined, they perform slightly better than when used individually. However, their predictive capabilities remain modest. While these features provide valuable information, they do not achieve the discriminative power of the model's predicted probabilities, which largely encapsulate the relevant predictive information. This finding suggests that additional factors - or more complex relationships captured by the model generating the predicted probabilities - play a more significant role in determining outcomes. In any case, it is important to note that the test set comprises 93 patients, of whom 64 are over 60 years old and 29 are under 60 years old. This distribution should be considered when interpreting the results.

\setcounter{figure}{2} % Optional: only if you need to force figure number
\begin{figure}[htbp!]
    \centering

    \begin{subfigure}[b]{\textwidth}
        \centering
        \includegraphics[width=\linewidth]{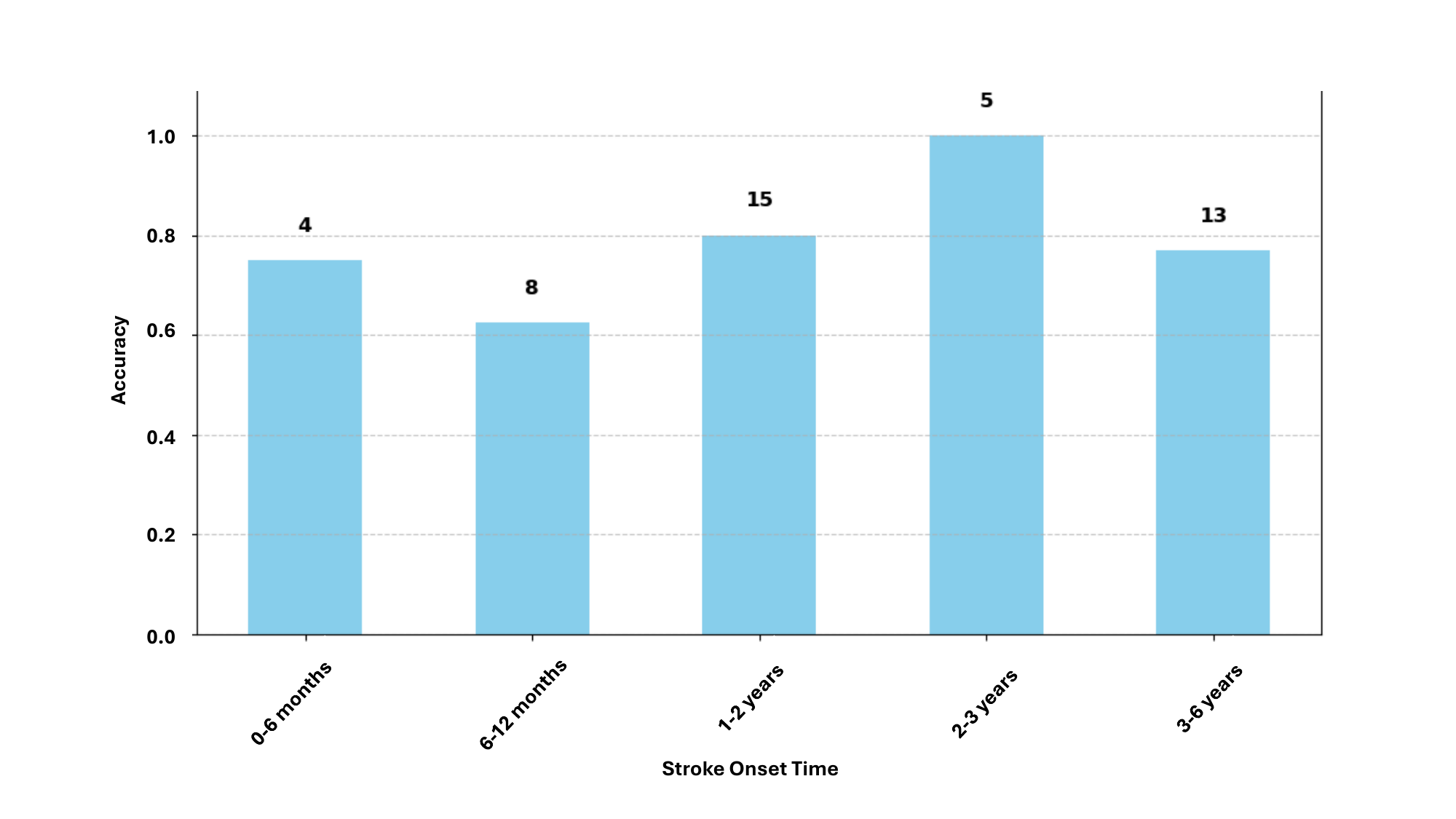}
        \caption{Model performance vs stroke onset time.}
        \label{fig:stroke_onset_time}
    \end{subfigure}
    
    \vspace{0.5cm} % Optional spacing between subfigures
    
    \begin{subfigure}[b]{\textwidth}
        \centering
        \includegraphics[width=\linewidth]{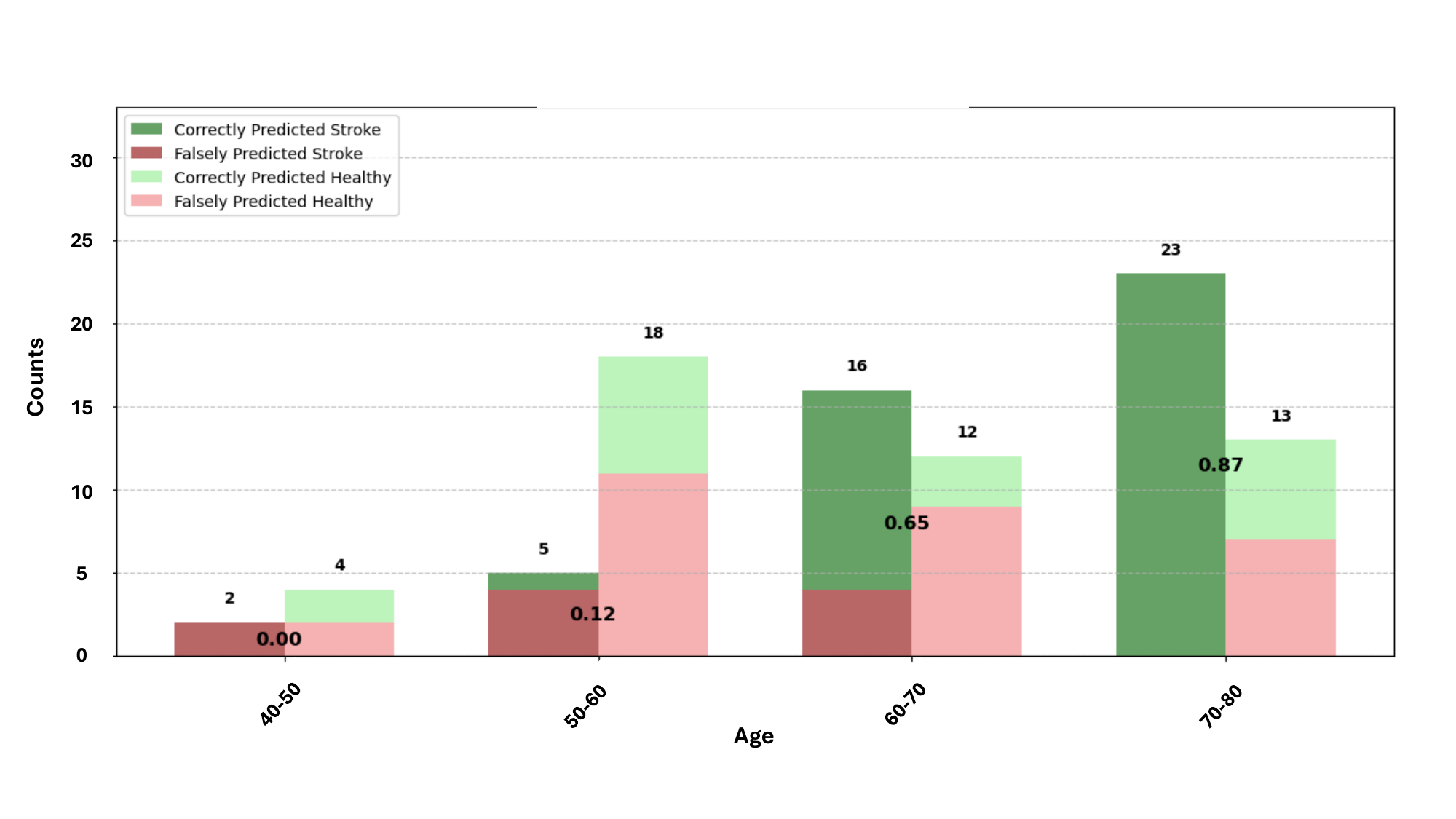}
        \caption{Model performance vs age.}
        \label{fig:age}
    \end{subfigure}
     
\caption{Distribution of metrics with respect to group partition. \textbf{(A)} Model accuracy with respect to stroke onset time for stroke patients. Number of patients in each group are given on top of bars. Stroke onset time is the time from scan acquisition to stroke onset. Note: intervals are left-inclusive and right-exclusive. \textbf{(B)} Distribution of the predictions with respect to the age group. Above the bars, the number of patients in each group is given and in the middle of each bar, the F1 score.}
\label{fig:subfigure}
\end{figure}

\subsection{Interpretability and qualitative analysis}
\label{subsec:interpretability_results}
\subsubsection{Embeddings visualization}
\label{subsubsec:emb_viz}
\begin{figure}[htbp!]
  \centering
  \includegraphics[width=\textwidth]{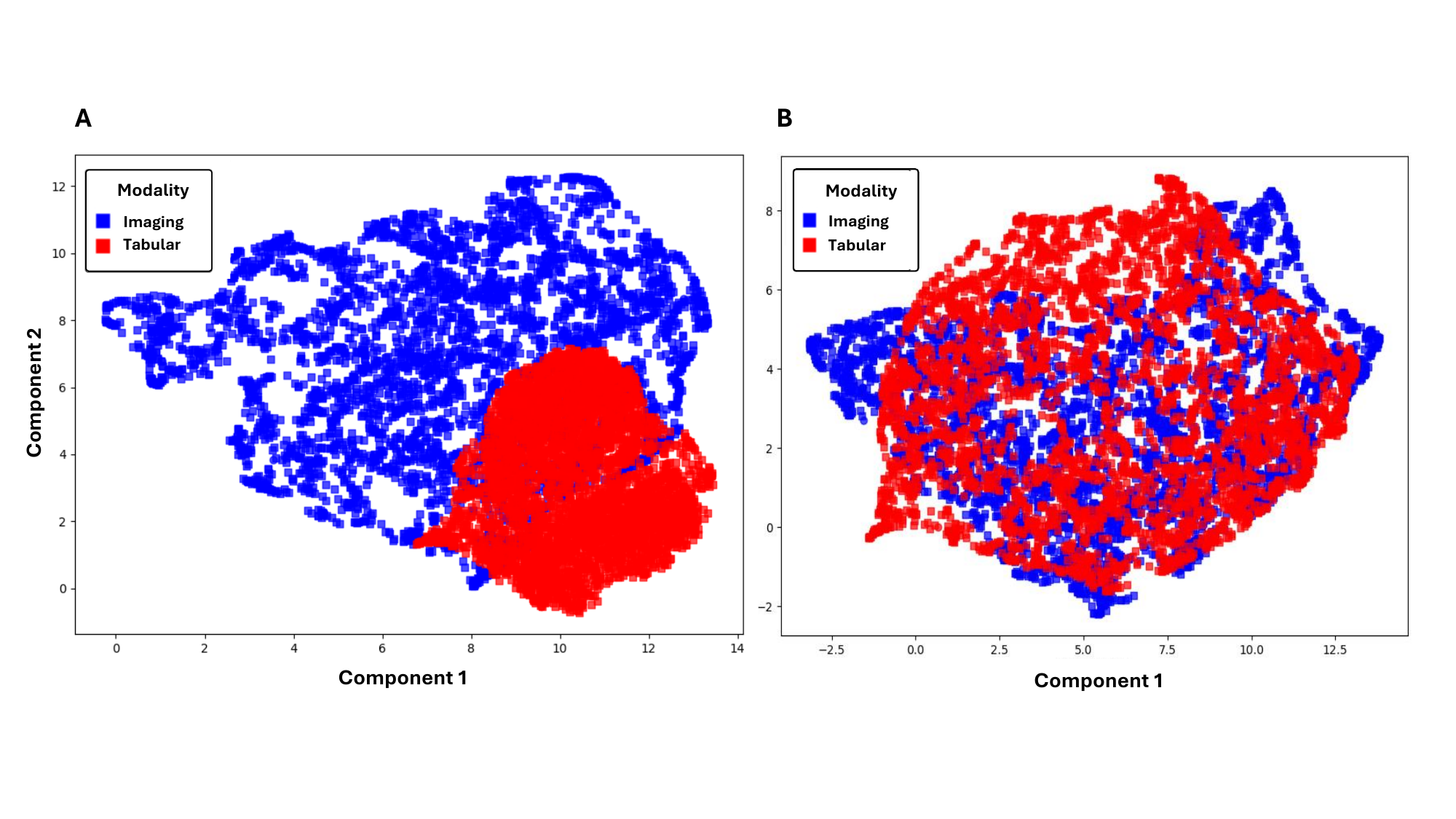}
   \caption{UMAP projections of tabular and imaging embeddings from validation set, using (A) uni-modal pre-trained tabular and imaging encoders and (B) multi-modal pre-trained tabular and imaging encoders.}
   \label{fig:umap_multi-modal}
\end{figure}

Figure \ref{fig:umap_multi-modal} shows the UMAP embeddings distribution for uni-modal and multi-modal data models. On the one hand, it can be observed that in Fig. \ref{fig:umap_multi-modal} A,  there is a clear distinction between (uni-modal learnt) tabular and imaging data modalities, with data samples clustered by data-type. In this case, the embeddings generated from imaging data and tabular data are different from each other in the feature space when generated with a uni-modal pre-trained model (i.e., either SCARF or SimCLR). The tight clustering of red points suggests that the tabular data embeddings are more homogeneous and possibly more concentrated in the feature space compared to the broad representation of brain MRI images. Therefore, uni-modal data encoders have learned modality-specific features, without capturing interactions between them. On the other hand, in Fig. \ref{fig:umap_multi-modal} B the UMAP plot obtained using our method (multi-modal, using CLIP and ITM). In this case, there is a consequent overlap between the tabular and imaging embeddings, suggesting that the model has found common representations for the two different data types, either via shared visual features or via learning associated clinical patterns in tabular and brain MRIs. Thus, our method is able to encode the underlying patient representation in a common latent space by reducing data augmentation noise. Still, there are data-points in the plot having distinct representations within each modality, suggesting that the model could not project them to the modality-shared latent space. The broad distribution of points across the entire UMAP space suggests that the embeddings capture a wide variety of features from both imaging and tabular data, rather than collapsing all data points into a narrow cluster. These results expose the enhanced performance of the multi-modal SSL strategy by projecting diverse data modalities into a shared embedding space, and thus suggesting a better model starting point for downstream analysis. 

\subsubsection{Imaging heatmaps and tabular feature importance}
\begin{figure}[p]
    \centering

    % Subfigure A
    \begin{subfigure}[b]{\textwidth}
        \centering
        \includegraphics[width=\textwidth, trim=0cm 0cm 10cm 0cm, clip]{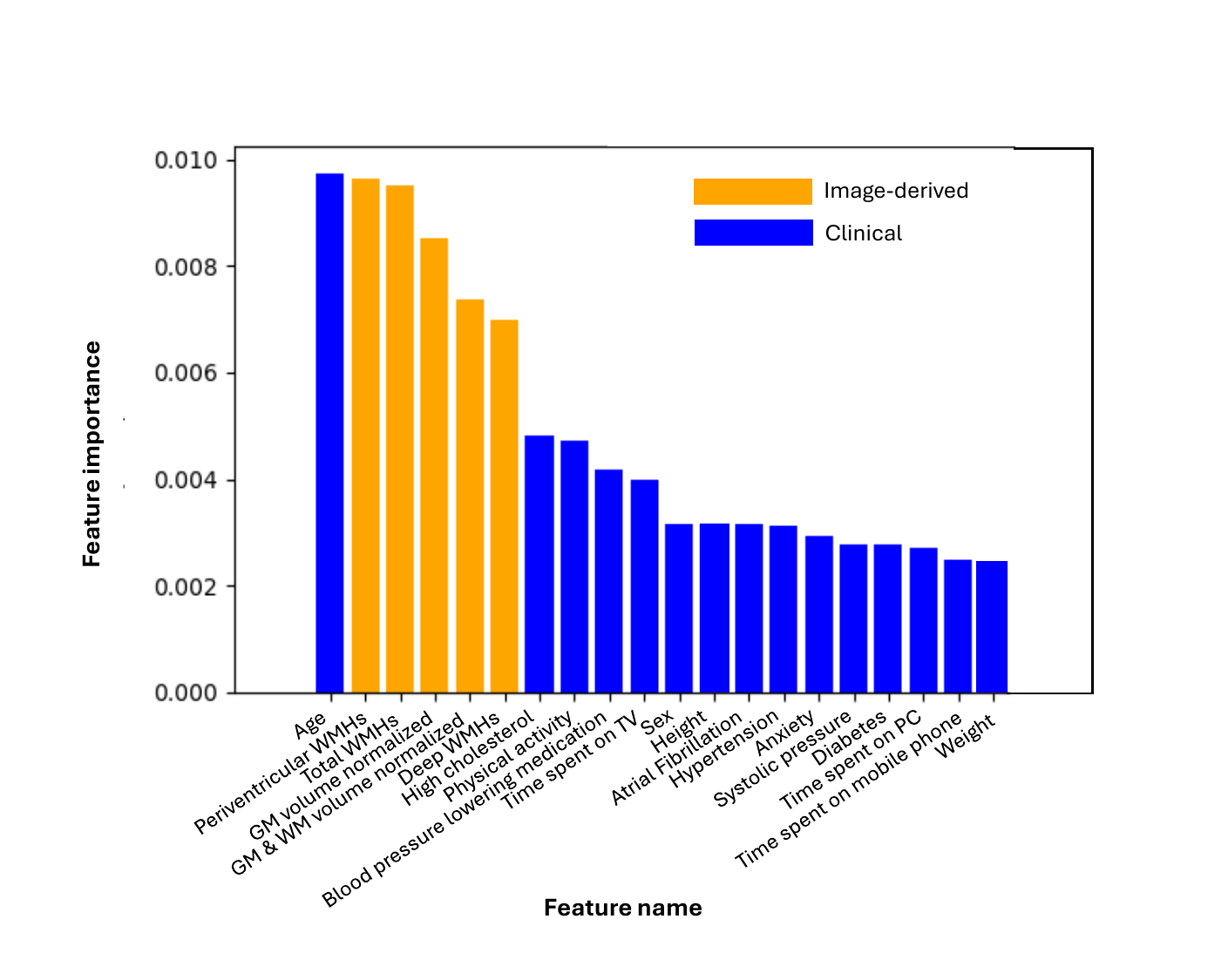}
        \caption{Tabular feature importance.}
        \label{fig:tab_importance}
    \end{subfigure}

    \vspace{1cm} % Optional space between figures

    % Subfigure B
    \begin{subfigure}[b]{\textwidth}
        \centering
        \includegraphics[width=\textwidth]{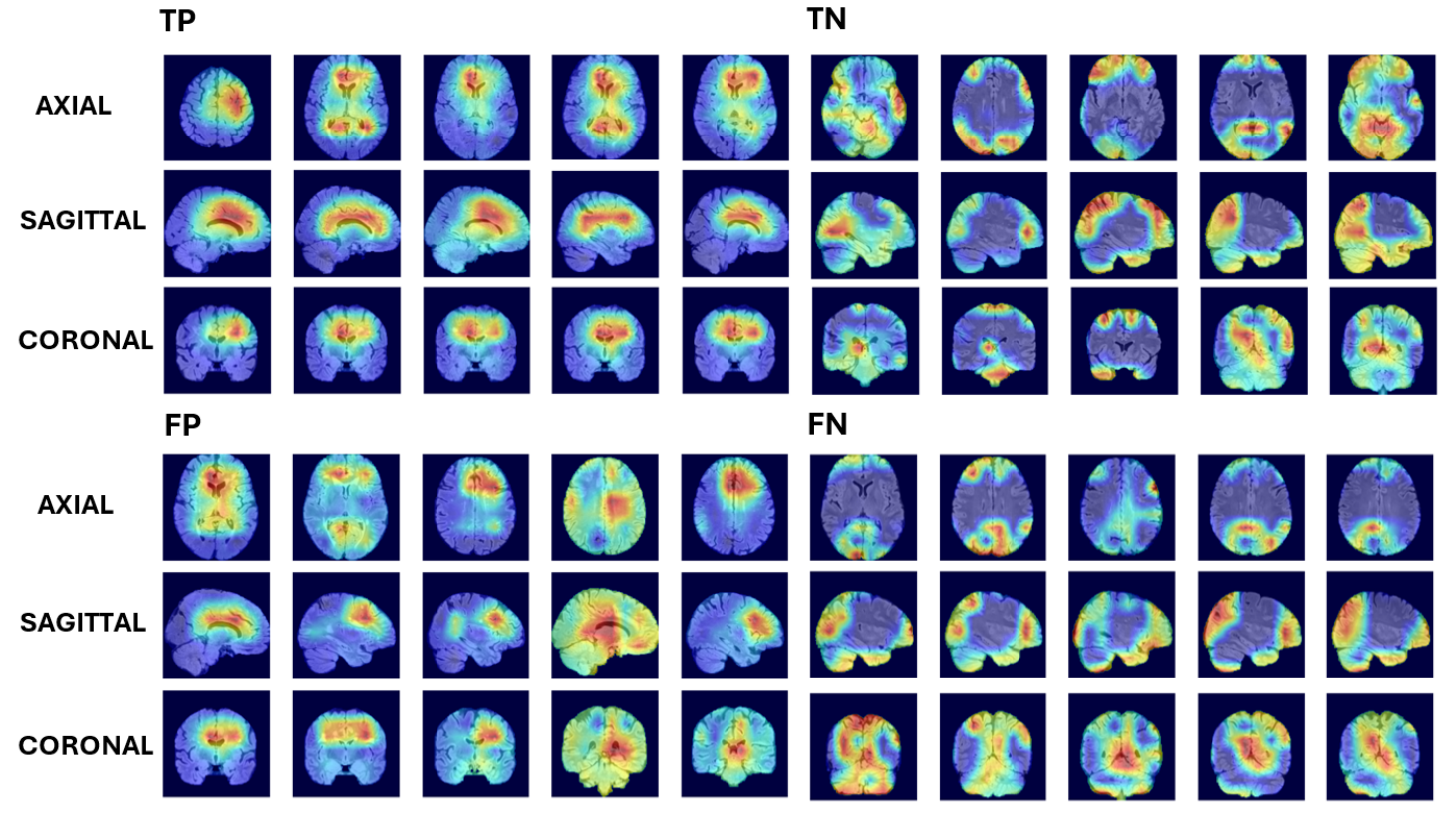}
        \caption{GradCAM-activated brain regions}
        \label{fig:GradCAM}
    \end{subfigure}

\caption{Feature importance analysis for model interpretability with anatomical T2-FLAIR scans and clinical records. \textbf{(A)} Tabular feature importance for image-derived (orange) and clinical features (blue) from integrated gradients analysis. \textbf{(B)} GradCAM-activated brain regions for five patients, categorized as TP (True Positive), TN (True Negative), FP (False Positive), and FN (False Negative). Red (blue) indicates higher (lower) activations.}
\label{fig:subfigure2}
\end{figure}

The analysis of tabular features importance, shown in Figure \ref{fig:tab_importance} demonstrates that age and image-derived features, such as PWMHs and DWMHs, contribute significantly to the predictions in the multi-modal model, likely indicating their robust representation of brain pathology. Clinical features like physical activity, cholesterol level, blood pressure lowering medication, hypertension, systolic pressure, previous symptoms of atrial fibrillation and sex also play key roles, though their relative importance is lower compared to the imaging-derived features. The use of integrated gradients allows for a nuanced understanding of the contributions from both modalities, highlighting how brain MRI features complement clinical records in predicting outcomes. This balanced contribution supports the potential of multi-modal models to integrate complex, heterogeneous data sources effectively for robust clinical insights. However, unlike the variable "Age", showing a statistically significant difference between groups in \ref{tab:demographics}, and is a key contributor to the model’s decision-making (Figure \ref{fig:tab_importance}), the variable "BMI", although also statistically significant, is not among the most influential tabular features (Figure \ref{fig:tab_importance}). The observed differences could still introduce subtle biases if the model were to generalize to populations with different BMI distributions. We acknowledge this as a potential limitation and suggest that future work could explore domain adaptation or covariate adjustment techniques to further mitigate residual confounding \cite{Austin2011PropensityScore}.

\label{subsubsec:heatmaps}
Figure \ref{fig:GradCAM} shows the results from the GradCAM experiment obtained over predicted samples. When inspecting the positive predicted scans (True Positives and False Positives), the model tends to highlight anatomical regions surrounding the lateral ventricles and (periventricular) white matter areas. Such patterns could be associated to WMHs, which are known predictors of brain atrophy and age-related brain alterations \cite{Grajauskas2019} and also stroke risk predictors in elderly individuals \cite{park2019white}. In different studies, correlations have been observed between common age-related structural brain changes and brain pathologies \cite{Grajauskas2019, Guo2017, Shim2015}.
In the True Positive patients class, the views of scans \#2 and \#4 in Fig. \ref{fig:GradCAM} with overlaid activation maps are highlighting distant anatomical regions from the lateral ventricles. Supported from literature, those activations could be related to deep WMHs, often appearing in brain regions that are not immediately adjacent to the cortical surface, but commonly located in subcortical white matter or in deep white matter tracts \cite{Grajauskas2019}. Such DWMHs are associated with chronic vascular disease and other chronic pathologies (e.g. multiple sclerosis) \cite{Grajauskas2019}. When looking at negatively predicted patients (True Negatives and False Negatives), the scans are putting less emphasis on the periventricular white matter region but instead highlight areas of the lower brain (cerebellum, posterior brain) and the cortex. We hypothesize that these areas may reflect patterns related to normal aging or normal brain atrophy processes, rather than anomalous brain conditions. Overall, we can hypothesize from these visualizations that the multi-modal SSL stroke risk predictor model focuses on pathological lesions for its predictions. We therefore believe that further experiments including brain age and brain structural age biomarkers could help enhancing the model's predictability, since they have been shown to be associated with overall cardiovascular risk \cite{de2020multimodal}, clinical outcome in stroke \cite{liew2023association} and overall risk of mortality \cite{cole2018brain}.

\begin{figure}[htbp!]
  \centering
  \includegraphics[width=0.6\textwidth]{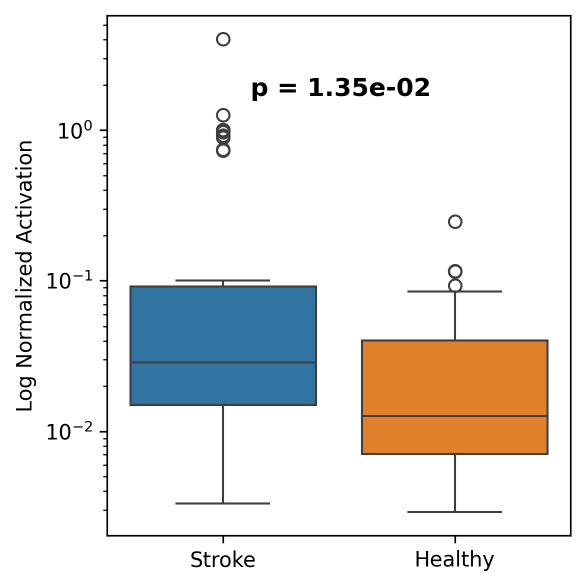}
  \caption{Boxplot of normalized GradCAM activations of white matter hyperintensities masks over whole brain. The distributions are constructed using per-scan average activation values. $p$-value is shown in bold.}
  \label{fig:boxplot}
\end{figure}

Figure \ref{fig:boxplot} shows the distributions of normalized average GradCAM activations across WMHs masks over the whole brains. As evidenced by the figure, stroke patients are showing a significantly higher mean activation in WMH regions than for overall brains compared to the healthy group, suggesting that GradCAM has identified more pronounced activations in regions correlating with stroke development \cite{Grajauskas2019}. A Mann-Whitney U test was performed to compare the normalized mean GradCAM activations between stroke and healthy groups and shows a statistically significant difference (U = 1403, $p$-value: 1.35 x 10$^{-2}$), indicating distinct normalized mean activation levels between the two groups.
As these activations highlight areas of interest, particularly in WMH lesions, it is likely that the model is focusing on relevant pathological regions when identifying stroke-related features. This significant difference underscores the model’s sensitivity to stroke markers, especially when WMH lesions are isolated from the whole brain MRI, indicating reliable discrimination between stroke and healthy brain images based on observed activations.

\textbf{Limitations.} Our study is limited by the use of the UK Biobank, whose demographic characteristics may not fully represent the diversity of global populations, potentially impacting the model’s generalizability and clinical utility. Future research should validate our approach using more diverse external datasets to improve applicability. Additionally, our test set was constrained by the limited availability of pre-stroke imaging samples, as most stroke datasets focus on post-onset cases. Therefore, it is not trivial to define a standardized time window where individuals would be monitored during the pre-stroke onset period. However, from the scan acquisition to either the stroke onset or the second healthy scan acquisition, our algorithm spans between 26 and 29 months on average for all participants.
Moreover, there is some variability in the model performance with respect to the time between imaging and stroke onset, even though there is no clear trend showing that it is negatively affecting the model performance. Future work could also include improving model efficiency by testing further architectures and techniques such as network pruning \cite{Wang2024} to reduce model parameters. Finally, analyzing saliency maps (e.g., Grad-CAMs) using calibrated models or advanced techniques (such as Score-CAM or Opti-CAM \cite{Wang2020ScoreCAM, Zhang2024OptiCAM, Moller2020ReliableSaliency}) could yield more precise insights into imaging biomarkers and improve voxel-level localization of brain regions that most strongly influence the model’s decisions, ultimately enhancing our understanding of stroke risk predictors.

\section{Conclusion}
\label{sec:conclu}
We hereby present a multi-modal foundation model integrating diverse data modalities for predicting stroke risk. The model performance is compared against state-of-the-art (self-) supervised models employing both uni-modal and multi-modal data, including tabular and imaging datasets. A comprehensive set of experimental settings is utilized, encompassing different subgroupings of tabular features - such as clinical records, brain IDPs, and lesion IDPs - as well as various training regimes that combine pre-training and fine-tuning based on data modality. 

Our results demonstrate that the CLIP model on multi-modal data, combined with an ITM loss, outperforms single-modality alternatives. Our method surpasses the self-supervised tabular (image) data SCARF (SimCLR) model by 2.1\% (2.8\%) in ROC-AUC, and by 10.6\% (12.8\%) in balanced accuracy terms. It also demonstrated a 0.93\% AUROC and 7.6\% accuracy improvements when compared with the best multi-modal supervised method. Additionally, the proposed model produces well-aligned multi-modal representations in a common, data modality-independent space, which is unattainable with uni-modal tabular or imaging data models. Thus, our method effectively leverages complementary and synergistic information from diverse data modalities.

Using interpretable heatmaps, we identified activated brain regions commonly associated with brain aging, stroke risk, and clinical outcomes. On one hand, the activated areas indicate that the model primarily focuses on deep and periventricular WMHs for predicting positive samples, which may be more common and extensive in patients identified as at risk for stroke. On the other hand, the prediction of negative samples highlights the cerebellum, posterior brain regions and cortical areas. These results demonstrate the model capacity to extract task-specific features linked to stroke risk, which are well-supported by existing literature. 

In conclusion, we propose a robust multi-modal foundation model for stroke risk prediction. Our model lays the groundwork for future studies aiming to integrate multiple data modalities into prediction models. 

\newpage

%%% Please be aware that for original research articles we only permit a combined number of 15 figures and tables, one figure with multiple subfigures will count as only one figure.
%%% Use this if adding the figures directly in the manuscript, if so, please remember to also upload the files when submitting your article
%%% There is no need for adding the file termination, as long as you indicate where the file is saved. In the examples below the files (logo1.eps and logos.eps) are in the Frontiers LaTeX folder
%%% If using *.tif files convert them to .jpg or .png
%%%  NB logo1.eps is required in the path in order to correctly compile front page header %%%

\section{Conflict of Interest Statement}
%All financial, commercial or other relationships that might be perceived by the academic community as representing a potential conflict of interest must be disclosed. If no such relationship exists, authors will be asked to confirm the following statement: 

The authors declare that the research was conducted in the absence of any commercial or financial relationships that could be construed as a potential conflict of interest.

\section{Non-standard abbreviations}

\begin{itemize}

    \item PWMHs, DWMHs, WMHs: Periventricular, Deep, white matter hyperintensities
    
    \item SSL: Self-supervised learning
    
    \item MRI: Magnetic Resonance Imaging
    
    \item CLIP: Contrastive Language-Image Pre-training
    
    \item ITM: Image-Tabular Matching
    
    \item CNN: Convolutional Neural Networks
    
    \item UKB: UK Biobank 
    
    \item ICD: International Classification of Disease
    
    \item IDPs: Image-Derived Phenotypes 
    
    \item MLP: Multi-Layer Perceptron
    
    \item UMAP: Uniform Manifold Approximation and Projection 
    
    \item GradCAM: Gradient-weighted Class Activation Mapping
    
    \item ROC: Receiver Operating Characteristic
    
    \item AUC: Area Under the Curve
    
\end{itemize}

\section{Funding Statement and Acknowledgments} 
This research has been conducted using the UK Biobank resource under application number 81959 and grants NHLBI R21HL156174, K24HL136852, R21HL167173. The project was supported by the grant \#2023-N-306 of the 1st Joint Call of the Swiss Data Science Center (SDSC) and the Strategic Focus Area “Personalized Health and Related Technologies (PHRT)” of the ETH Domain (Swiss Federal Institutes of Technology) and DataSpectrum4CVD and the grant from PHRT Project \#2022-812 / SDSC C22-15P. BM, ND, and EDLR acknowledge support of the Helmut-Horten-Foundation.

\section{Data Availability Statement}
The UK Biobank dataset is semi-private. You can apply for an academic access here: \href{https://www.ukbiobank.ac.uk/enable-your-research/apply-for-access}{access UKBB}. The code repository for this project is available at the following link: \href{https://github.com/CamilleDelgrange/SSMSRPM/}{SSMSRPM}. 
% Please see the availability of data guidelines for more information, at https://www.frontiersin.org/about/author-guidelines#AvailabilityofData

%\section*{References}
% Bibliography section
\newpage
\printbibliography  % Name of your .bib file 

\end{document}